\documentclass{svproc}
\usepackage{url}
\usepackage{graphicx}
\usepackage[caption=false]{subfig}
\usepackage{hyperref}
\usepackage{xcolor}

\begin{document}
\mainmatter              
\title{A Toolkit to Generate Social Navigation Datasets}
\titlerunning{A Toolkit to Generate Social Navigation Datasets}  
%

\author{
Rishabh Baghel\thanks{These authors contributed equally to this work.} \inst{1} \and Aditya Kapoor$^\star$\inst{2} \and Pilar~Bachiller\inst{3} \and Ronit~R.~Jorvekar\inst{4} \and Daniel~Rodriguez-Criado\inst{5} \and Luis~J.~Manso\inst{5}
}
\authorrunning{Baghel et al.} 
%
%
\institute{
Dept. of Computer Science and Engineering,\\Indian Institute of Information Technology Guwahati, India \\
\and
Dept. of Computer Science and Information Systems,\\Birla Institute of Technology and Science Goa, India\\
\and
Robotics and Artificial Vision Laboratory, University of Extremadura, Spain
\and
Dept. of Computer Engineering, Pune Institute of Computer Technology, India
\and
Dept. of Computer Science, College of Engineering and Physical Sciences,\\ Aston University, United Kingdom,\\
\email{l.manso@aston.ac.uk}\\
}

\maketitle              

\begin{abstract} 
Social navigation datasets are necessary to assess social navigation algorithms and train machine learning algorithms.
Most of the currently available datasets target pedestrians' movements as a pattern to be replicated by robots.
It can be argued that one of the main reasons for this to happen is that compiling datasets where real robots are manually controlled, as they would be expected to behave when moving, is a very resource-intensive task.
Another aspect that is often missing in datasets is symbolic information that could be relevant, such as human activities, relationships or interactions.
Unfortunately, the available datasets targeting robots and supporting symbolic information are restricted to static scenes.
This paper argues that simulation can be used to gather social navigation data in an effective and cost-efficient way and presents a toolkit for this purpose.
A use case studying the application of graph neural networks to create learned control policies using supervised learning is presented as an example of how it can be used.
\end{abstract}
\section{Introduction}
To boost robots' prevalence in the service industry, health and home settings, their efficiency and social acceptance must be improved.
This lack of social acceptance requires a seamless interaction with humans that cannot be achieved without considering humans' goals, emotions, and predicting their future behaviour.
These skills are essential to successfully work with humans and avoid user rejection~\cite{de2015sharing}.
\par

Social acceptance is particularly important for mobile robots.
For instance, while moving, robots should estimate human trajectories to avoid getting in their way.
Robots should gather information about the subjects' current activity, emotions, and intentions to provide an efficient and natural interaction while offering them help. 
Let us consider a scenario where a guest arrives at the robot's owner's apartment, and they start having a heated argument just after that.
Should the robot approach the guest to offer a cup of tea as it would usually do?
Approaching humans in those conditions is probably something that roboticists would like to avoid.
The problem in this scenario is that the robot might not be able to interpret the social context correctly and assess which actions are acceptable.
The general challenge while programming robots with some degree of ability to understand humans is to cope with the complexity and breadth of factors involved. 
\par

The algorithms in charge of controlling robots and those estimating humans' emotions and aims are usually manually programmed, but they can also be based on Machine Learning (ML) models or a combination of both approaches.
Handcrafted algorithms require a considerable amount of time and are hard to debug, expensive to develop, and even expert developers might disregard variables that could be of use.
ML models are very often able to consider factors that go beyond experts' intuition.
However, until recently, we lacked ML algorithms to work with structured data such as graphs, so most ML-based algorithms tend to discard structural information or make poor use of it.
Recent non-Euclidean ML techniques such as Graph Neural Networks (GNNs)~\cite{Battaglia2018} can work with arbitrarily complex structured data in the form of graphs, as opposed to the Euclidean structures that conventional ML approaches are limited to.
\par

In social navigation, datasets are crucial to a) assess algorithms' appropriateness and; b) to train ML models for social navigation.
Although this holds regardless of whether structured or conventional ML models are used, the toolkit presented in this paper focuses on -but is not limited to- structured (\textit{i.e.}, graph-like) data.
The motivation is that structured data can naturally be represented as a graph~\cite{Battaglia2018}, but the opposite does not always hold.
\par

Given the subjective and highly-interactive nature of social navigation interactions and Human-Robot Interaction (HRI), generating large datasets for social navigation in real-life scenarios would be prohibitively expensive and time-consuming.
Previous works have used a tool to generate a social navigation dataset, SocNav1.
The dataset, presented in~\cite{manso2020socnav}, was designed to be used to train and benchmark models to assess the social compliance of robots in a given scenario.
Although SocNav1 is limited in many aspects (covered in the next section), it is one of the few datasets considering not only metric properties but also relationships, which can be of high importance and most datasets neglect.
\par

The contributions of the paper at hand are a new open-source modular toolkit based on CoppeliaSim and PyRep~\cite{rohmer2013coppeliasim,james2019pyrep} to generate social navigation datasets, and a use-case where the toolkit is applied to create a dataset.
The tool aims to enable third-party researchers to develop their datasets while reducing the time spent developing the required tools.
The use-case provided demonstrates that the data generated using the toolkit can be used to train robots in social navigation tasks.

\section{Navigation datasets}
There are many publicly available datasets and tools that can be used for social navigation.
These datasets can be classified as video recordings, geometric data, symbolic data, audio, or multimodal data -or a combination of them.
Video datasets, which are the most common type, can be divided into real or simulated.
EIPD~\cite{EIPD}, a video dataset, was used to learn the behaviour of pedestrians in~\cite{LuberSociallyAwarerobot}.
The video datasets ETH~\cite{ETH} and UCY~\cite{UCY} were used in~\cite{socialattentionn} to train a social attention model.
ETH contains two sets of videos that were obtained from a bird's eye view and were manually annotated.
It has a total of $650$ human trajectory tracks of over $25$ minutes.
A dataset for public space surveillance tasks was also presented in~\cite{PETS}.
It consists of $28$ videos from $6$ in different scenarios.
Another dataset of videos from CCTV cameras involving pedestrians is presented in~\cite{benfold2009attention}.
In addition to pedestrians, \cite{Robicquet2016SDD} contains videos of other agents in a social scenario such as bicyclists, cars, buses, and even skateboarders.
Hence, \cite{EIPD,ETH,UCY,PETS,benfold2009attention,Robicquet2016SDD} belong to the category of video datasets.
Recording videos is relatively easy and inexpensive, but the videos need to be annotated, and even if there is some form of person detector to generate the annotations, these need to be supervised by humans.
Similarly, videos have no semantic features \textit{per se} (\textit{e.g.}, who is talking to whom), so this kind of information is missing or has to be added by hand.
\par

The Carnegie Mellon University (CMU) dataset~\cite{CMUDataset}
is a 3D dataset which was used in~\cite{socialmapping} to encode the context-dependent spatio-temporal interactions into social-maps.
This dataset contains $2605$ pre-recorded action sequences, and $55$ of these sequences correspond to social interactions between two subjects.
It involves two subjects, whereas in real-life, the social scenarios contain more entities and are more complex in nature.
A multimodal dataset with $64$ minutes of multimodal sensor data is presented in~\cite{martnmartn2019jrdb}.
It includes omnidirectional and regular RGB video streams, 3D point cloud data from a LIDAR sensor, audio signals, and encoder values from a robot's wheels.
\par

Simulators are also a useful tool while collecting data as they are cost-effective, and because the state of the world is fully observable, there is no need for manual annotations. 
On the other hand, one of the main drawbacks of simulations is that they will always yield approximations to real data, and that might introduce biases.
Pedestrian simulations were used in~\cite{Ferrer2013} to fit a social force model where the model parameters were learned using thousands of simulations.
The UCY dataset~\cite{UCY}, which was mentioned in video datasets, could also fall in this category because it introduces a data-driven simulation approach where agents' behaviour is based on thousands of previous examples in the database.
The SocNav1 dataset~\cite{manso2020socnav} combines geometric and semantic relations.
Instead of containing subjects moving, SocNav1 provides a series of scenes with humans and a robot.
The labelled information corresponds to the \textit{social score} of the robot in that given position.
One of the advantages of SocNav1 is that it combines geometric with symbolic information, so social relations can be considered to estimate the score.
The dataset's main limitation is that the scenes are static, so it does not consider human movement.
\par

From the reviewed datasets, only SocNav1 provides information related to how humans want robots to behave.
Instead, most datasets provide human movement information, expecting that robots should mimic their behaviour.
Mimicking human navigation is not a bad alternative but in some cases robots are expected to behave differently.
For instance, it could be interesting to make robots especially careful regarding social distances (\textit{e.g., "when moving around older people or people handling fragile objects"}), or make them extra cautious when it comes to causing interruptions (\textit{e.g., "be extra careful not to go in the way of people walking or having a conversation"}).
However, SocNav1 is limited to static scenes, which neglects the importance of the robots' trajectories and the people around them.
The gap identified during the review is the lack of datasets targeting how people would like robots to move, which might differ from how pedestrians move.
The next section describes our proposal, whose aim is not to provide a dataset but a tool to create them according to the users' needs.

\section{Proposal}\label{proposal}
\begin{figure}[h]
    \centering
    \includegraphics[width=0.9\textwidth]{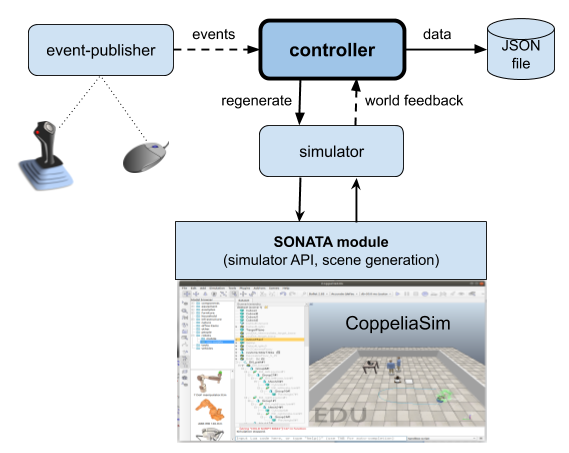}
    \caption{General structure of the proposed toolkit for dataset generation.}
    \label{fig:general_structure}
\end{figure}

No one-size-fits-all approach will suffice for generating datasets, as different datasets will work with different kinds of data.
Therefore, instead of creating an off-the-shelf solution, our SOcial NAvigation Toolkit for data Acquisition (SONATA) provides software components that can be used as scaffolding to implement different dataset generation tools.
The general structure of the toolkit is shown in Fig.~\ref{fig:general_structure}.
It is composed of three software components (rounded squares) and a Python module.
The module is built on top of PyRep~\cite{james2019pyrep}, which is in turn used to communicate with CoppeliaSim~\cite{rohmer2013coppeliasim}.
The module's goal is twofold: a) to provide a high-level API to create manual or randomly-generated social navigation scenes, and b) to control the simulation and record data.
\par

The toolkit architecture has been designed as a network of components to reduce software coupling and ensure that the computational load can be distributed among different cores or even different computers.
The \textit{simulator} component instantiates the \textit{CoppeliaSim} simulation, publishes the information about the simulated entities (people, walls, objects, and goals), and acts as an interface with the simulator, handling robot control and scene regeneration commands from a \textit{controller} component.
The \textit{event-publisher} component receives events from a joystick or mouse and publishes the corresponding data.
These two software elements are common to every specific dataset generation tool built using SONATA.
The only problem-specific component is the \textit{controller}, so the toolkit has been designed to facilitate forking the project and modifying the controller component.
\par

\begin{figure}[h]
    \centering
    \includegraphics[width=1.0\textwidth]{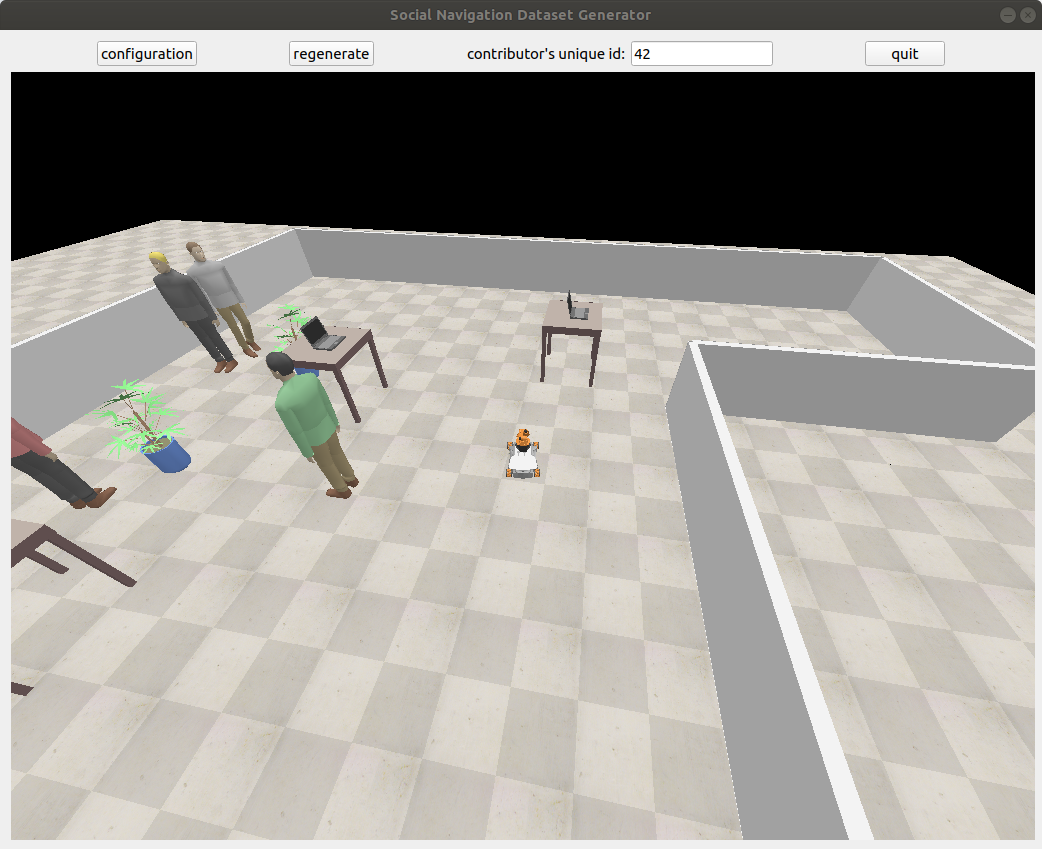}
    \caption{GUI example for the controller component.}
    \label{fig:screenshot}
\end{figure}

The \textit{controller} is subscribed to the event publisher and the simulator feedback.
It is therefore in charge of gathering data from the simulation and the user input, as well as controlling what happens in the simulation by sending commands to the \textit{simulator} component (\textit{e.g.}, moving the robot, the humans, restarting the simulation once a sample has been acquired).
Its default GUI contains a view of the simulation and options related to the generation of scenarios (see Fig.~\ref{fig:screenshot}).
The controller obtains data related to the simulation from the subscription to different topics provided by the \textit{simulator} component.
Specifically, in a given scene, the simulator publishes geometrical attributes of each entity included in it (\textit{e.g.}, humans, objects, walls) as well as semantic information about relationships between entities (\textit{e.g.}, two humans talking, two humans walking together).
Besides this information, the simulator component publishes an image of the scene captured by a virtual camera, which provides a third-person view of the robot.
This image, which is shown in the main GUI of the controller, provides a general view of the scene to help the user label it or provide feedback according to the specific problem. 
The controller also interacts with the simulation using two additional interfaces implemented in the \textit{simulator} component, which is accessible through proxy objects.
One of these interfaces provides an operation to regenerate the scenario.
For each type of entity (objects and relations), the user can set a minimum and maximum number of instances and randomly generate a new scenario according to the specified ranges.
The second interface is used to control the base of the simulated robot.
This control can be used as part of the labelling process or to change the user's point of view.
Fig.~\ref{fig:topics} summarises the topics that the controller component has access to.

\begin{figure}%
  \centering
  \subfloat[8cm][human sequence topic]{
   \begin{tabular}{|c|c|} \hline
    variable name & variable type \\ \hline
    id & integer \\
    x, y, angle & float  \\
    ix, iy, iangle & float \\
    \hline
  \end{tabular}}%
  \qquad
  \vspace{3mm}
  \subfloat[][wall sequence topic ]{
   \begin{tabular}{|c|c|} \hline
    variable name & variable type \\ \hline
    wall id & integer  \\
    x1, y1 & float  \\
    x2, y2 & float  \\
    \hline
  \end{tabular}}%
  \qquad
  \vspace{3mm}
  \subfloat[][goal topic]{
   \begin{tabular}{|c|c|} \hline
    variable name & variable type \\ \hline
    identifier & integer \\ 
    x, y & float  \\
    \hline
  \end{tabular}}%
  \qquad
  \subfloat[][joystick event topic]{
   \begin{tabular}{|c|c|} \hline
    variable name & variable type \\ \hline
    axis id & integer  \\
    value & float \\
    \hline
  \end{tabular}%
}%
  \qquad
  \subfloat[][object sequence topic]{
   \begin{tabular}{|c|c|} \hline
    variable name & variable type \\ \hline
    id & integer \\
    x, y, angle & float \\
    sideX, sideY & float \\
    \hline
  \end{tabular}  }
 \qquad
  \subfloat[][interactions topic]{
   \begin{tabular}{|c|c|} \hline
    variable name & variable type \\ \hline
    entity1 id & integer \\
    entity2 id & integer \\
    interaction type & string \\
    \hline
  \end{tabular}  }

  \caption{Topics provided by the simulator and the event-publisher components}%
  \label{fig:topics}%
\end{figure}

Along with the simulation data, the controller may access the user input to assign a label to each scene.
The user input reaches the controller through its subscription to a topic published by the \textit{event-publisher} component.
The received device events can be transformed into labels for a particular dataset.
Thus, to name some examples, an event can be associated to a binary (true/false) user response to a given question about the scene, a degree of acceptance of the user to a particular situation, or an action performed by the robot.
\par

The main control loop of the default \textit{controller} component periodically associates the current state of the scene with the label specified by the user and updates a list containing labelled data of the scenario.
The list is stored in a JSON file on user demand.
By default, the name of the file contains a user identifier and the timestamp to identify the file uniquely.
The choice of using JSON over other formats is motivated by its readability and ease to export to other formats.
Nevertheless, using other formats is possible and only depends on the developer's preferences for the dataset generation tool.

\section{Use case}\label{usecase}
A use case is presented to help readers understand how the toolkit is used and provide evidence of its usefulness.
The use case, which is accompanied by brief experimental results, covers the design and implementation of a dataset generation tool and how the data was gathered.

\subsection{Use case description}
The application of the dataset obtained from the use case is to train a supervised social navigation machine learning model to move a robot.
The particular model described later in this section is a Relational Graph Neural Network (R-GCN)~\cite{Schlichtkrull2018}.
The implementation details and the model's experimental results are not deeply described as the aim of the use case is only to provide an example of use.
However, the model serves the purpose of studying the limitations of supervised learning for robot control. 
\par
\begin{figure*}[!ht] 
 \centering
\subfloat[{Invading the people's personal space.}]{\includegraphics[width=0.48\textwidth,keepaspectratio=true]{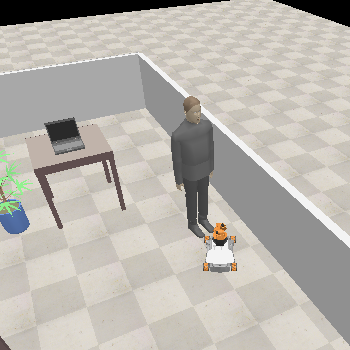} \label{fig:invading}}
 \hspace{0.1cm}
\subfloat[Getting in the way of people walking.]{ \includegraphics[width=0.48\textwidth,keepaspectratio=true]{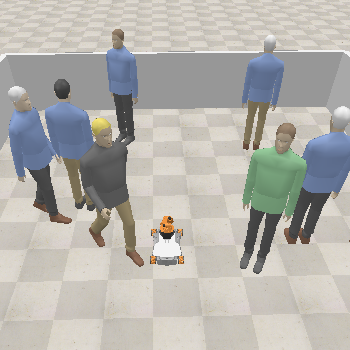}\label{fig:intheway}} \\
\subfloat[Interrupting people's interactions.]{\includegraphics[width=0.48\textwidth,keepaspectratio=true]{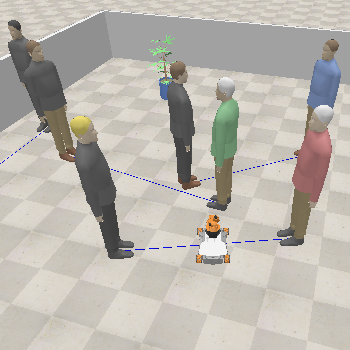} \label{fig:blocking}}
 \hspace{0.1cm}
\subfloat[Situation where the robot would have to move slowly.]{ \includegraphics[width=0.48\textwidth,keepaspectratio=true]{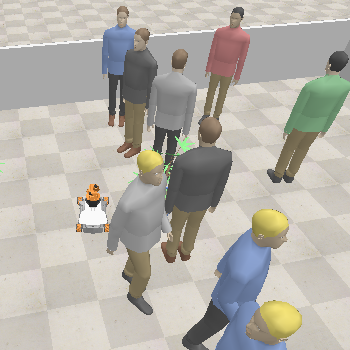}\label{fig:slowly}}
\caption{Examples of robot behaviour that should be avoided.}
\label{fig:nope}
\end{figure*}

Given that the model developed is used to control a robot using supervised learning, the model's output is the control commands that are meant to move the robot.
Thus, the desired outputs in the dataset are such control commands.
The input data are sequences of the different states of the environment where the robot is asked to move and the goal and the robot itself.
Given that the goal would be to achieve social navigation, the users providing the input-output relation (\textit{i.e.},~the sequence of robot commands for a given scenario) were asked to make the robot move towards the goal according to the following guidelines: a) avoid invading people's personal space; b) avoid getting in the way of people when they walk; c) avoid getting in the interaction area of people when they are interacting with an object (\textit{e.g.}, looking at a computer screen) or another person; and d) adjust the speed not to have a big impact on people's comfort.
Examples of these situations to avoid are shown in Fig.~\ref{fig:nope}.

\par

The dataset scenarios are rectangular and L-shaped rooms, with a randomised number of humans, tables, laptops, and plants (the numbers and types of objects do not have a significant impact on this task).
Some humans are static, while the rest randomly walk to places in the room.
Static humans might or might not be looking at one of the laptops in the scenario or talking to each other.
Moving humans might or might not be walking, accompanied by other fellow humans.
\par

The subjects contributing to the datasets have a third-person view of the robot and the environment, such as the one shown in Fig.~\ref{fig:screenshot}.

\subsection{Use case design and implementation}
As discussed in section~\ref{proposal}, developers only need to modify the \textit{controller} component.
It has to be adapted to perform two main tasks: a) save the data of interest, and b) control the simulation.
To make the above tasks as easy as possible, the controller is subscribed by default to the topics described in~\ref{proposal}, and the data is accessible as instance variables.
To control the simulation, the users can access operations available through different proxies charged with synchronous communication with the simulator component.
Every call to a method of a proxy invokes a remote operation in a remote component.
Thus, all the communication-related subjects are transparent for the developer.
\par

For every scenario in this specific use case, users are asked to move the robot following social conventions from a random location in the scene to a particular position using a joystick or mouse.
The controller of this use case is designed to interpret device events as velocity commands of the robot, but the joystick events could be used for any other purpose if necessary (\textit{e.g.}, provide rewards).
In this case, the controller sends movement commands to the simulated robot through the corresponding simulator's proxy.
Also, the command is also stored as the desired output for the current state of the scenario.
For each time step, the description of the world's current state and the robot command are appended to a list which is serialised into a JSON file at the end of each sequence (\textit{i.e.}, a JSON file is generated for each goal met).
Once the robot reaches the goal position, the controller stops the simulation, and the user decides whether or not the data should be kept or discarded.
After that, a new scenario is generated.
Additionally, the user can regenerate the scenario by using the corresponding GUI button.
\par

\subsection{Experimental results}
To test the data acquisition process, a set of $3092$ sampled trajectories ($6184$ including their corresponding mirrored versions) generated by the authors of the paper were stored, and an R-GCN was trained to control the robot.
The R-GCN has $5$ layers with $40$, $30$, $20$, $15$, and $3$ units, respectively.
The 3 units' output in the last layer is used to control the robot as advance, lateral, and rotation speed.
\par

The graphs used as input are composed of room representations for several consecutive time frames. 
This is done to let the network gather information about how the room's entities evolve through time.
Therefore, the creation of the final graph is done in three steps:
Firstly, graphs representing the room are created for two or three frames in a row. 
Secondly, these graphs are bonded, creating temporal edges from one node to the same one in the next time frame.
Finally, a feature vector is associated with each node adding valuable information for training the network.
\par

The graphs have $5$ different types of nodes, one for each element in the scenes: people, objects, walls, the goal, and the room node acting as a global node. 
Thus, there are bidirectional edges from the room node to every other node in the graph. 
Additionally, edges are also used for describing interactions between entities (\textit{e.g.}, a human talking to another human, a human interacting with an object).
All feature vectors have the same dimensions since it is a limitation of the network architecture. 
They are composed of different fields concatenated one after the other making a total of $42$ elements. 
The following equation shows the structure of the feature vector for node number $i$ in layer $0$ (the input layer):
$$h_i^{(0)} = (OH_{t_i} | OH_{f_i} | ts_i | p_i | o_i | r_i | w_i | g_i)$$
The first two fields ($OH_{t_i}$ and $OH_{f_i}$) are one-hot encodings.
The first one is for the type of nodes, so it has 5 elements (one for each type of node).
The second one encodes the time frame, which has a length of 3, given that the graphs are composed of a maximum of 3 frames.
$ts_i$ encodes the time passed since the beginning of the sequence, and has a dimension of $2$.
The remaining fields store normalised geometric information for each of the entities of the graph.
For instance, if the node is a person, the $p_i$ field will store its coordinates, velocity, orientation.
The rest of the geometric feature fields will be filled with zeros.
All the geometric features are computed from the robot's reference frame.
\par

The hyperparameters of the R-GCN were:
\begin{itemize}
    \item Number of layers: 5
    \item Hidden units: 50 (first hidden layer), 40, 30 and 15 (last hidden layer)
    \item Output units: 3 (velocity components of the robot)
    \item Activation functions: \textit{leaky ReLU} for the hidden layers and \textit{tanh} for the output layer
    \item Learning rate: 0.0005
    \item Weight decay: $10^{-9}$
    \item Batch size: 40
\end{itemize}
\par

\begin{figure*}[!ht] 
 \centering
\subfloat{\includegraphics[width=0.48\textwidth,keepaspectratio=true]{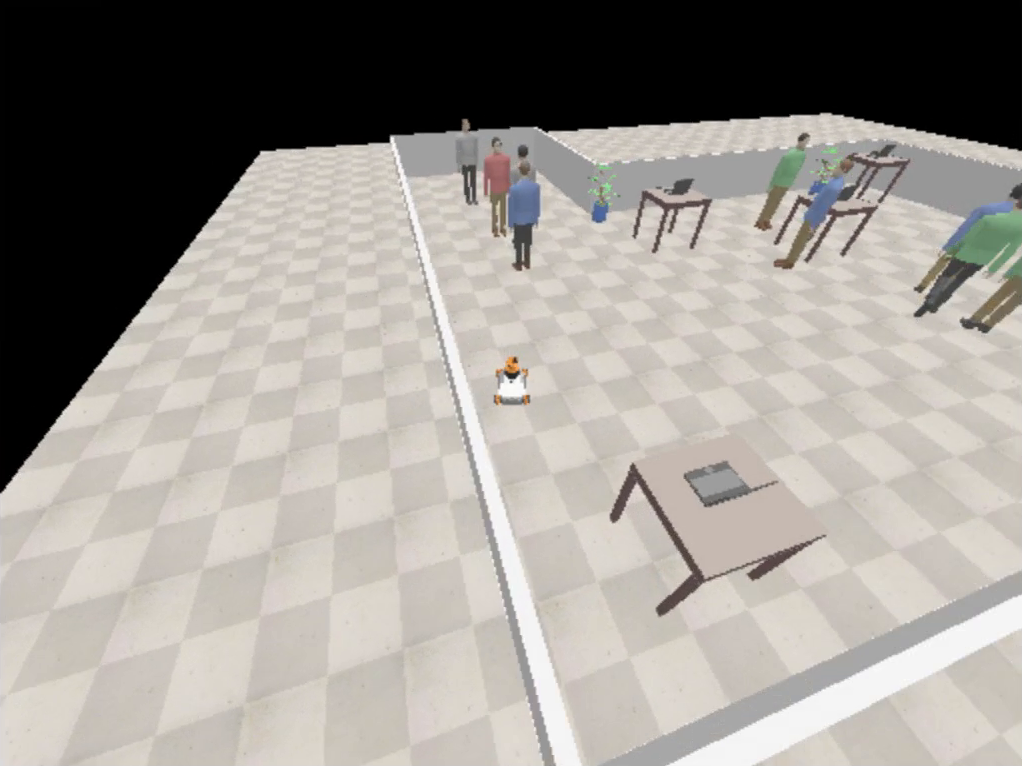}\label{fig:exp3_1}}
 \hspace{0.1cm}
\subfloat{\includegraphics[width=0.48\textwidth,keepaspectratio=true]{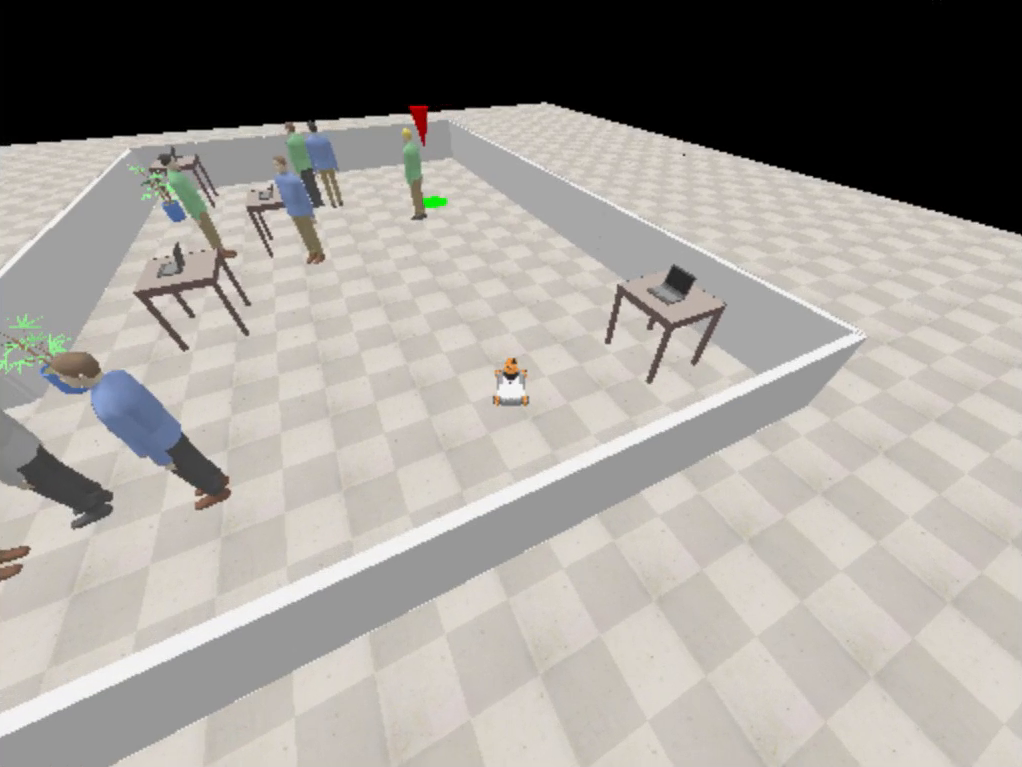}\label{fig:exp3_2}} \\
\subfloat{\includegraphics[width=0.48\textwidth,keepaspectratio=true]{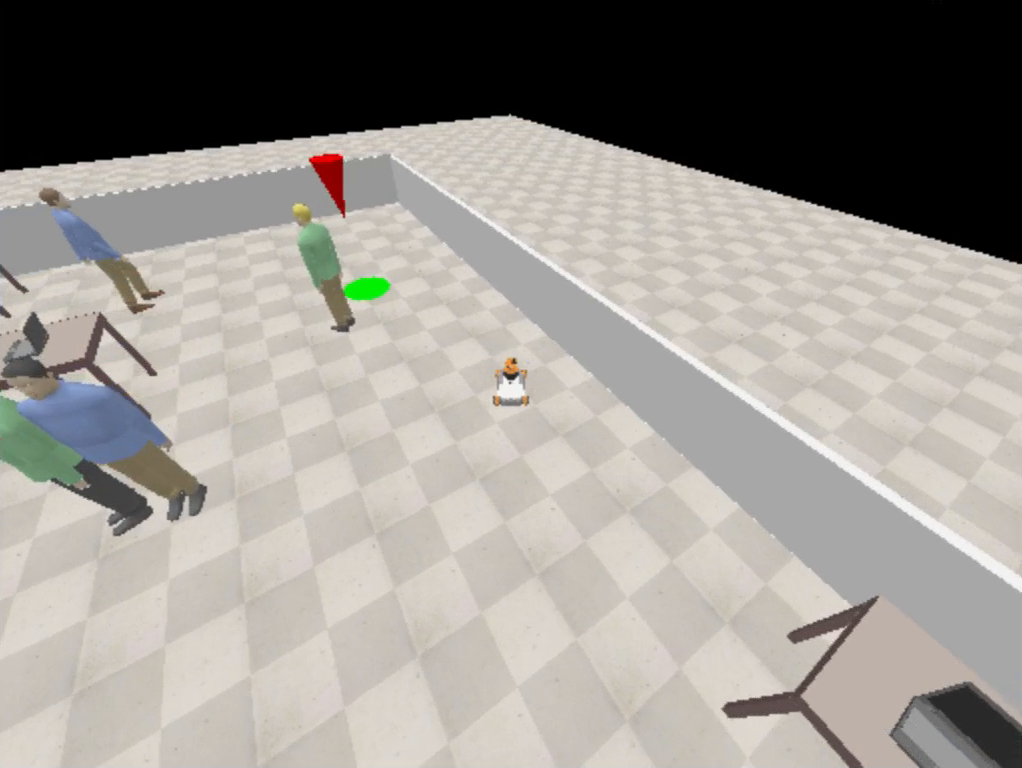}\label{fig:exp3_3}} 
 \hspace{0.1cm}
\subfloat{\includegraphics[width=0.48\textwidth,keepaspectratio=true]{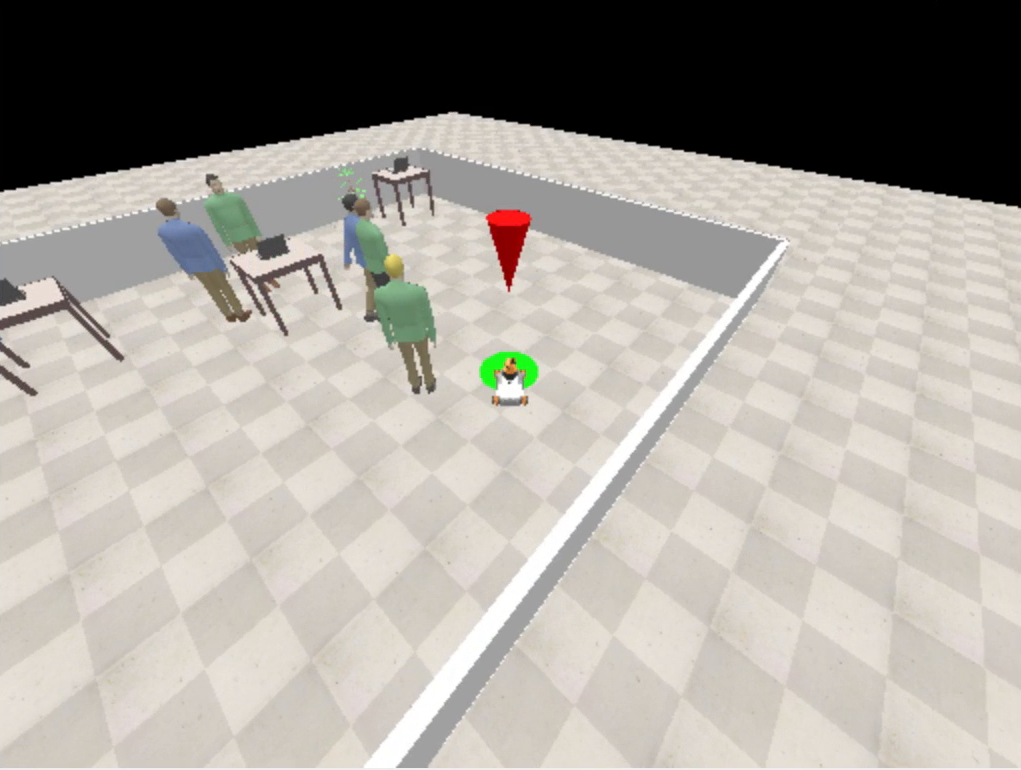}\label{fig:exp3_4}}
\caption{Execution of the robot using the trained model (first scenario). The robot correctly orients its body to avoid a standing human on the way to the goal position.}
\label{fig:exp3}
\end{figure*}

\begin{figure*}[!ht] 
 \centering
\subfloat{\includegraphics[width=0.48\textwidth,keepaspectratio=true]{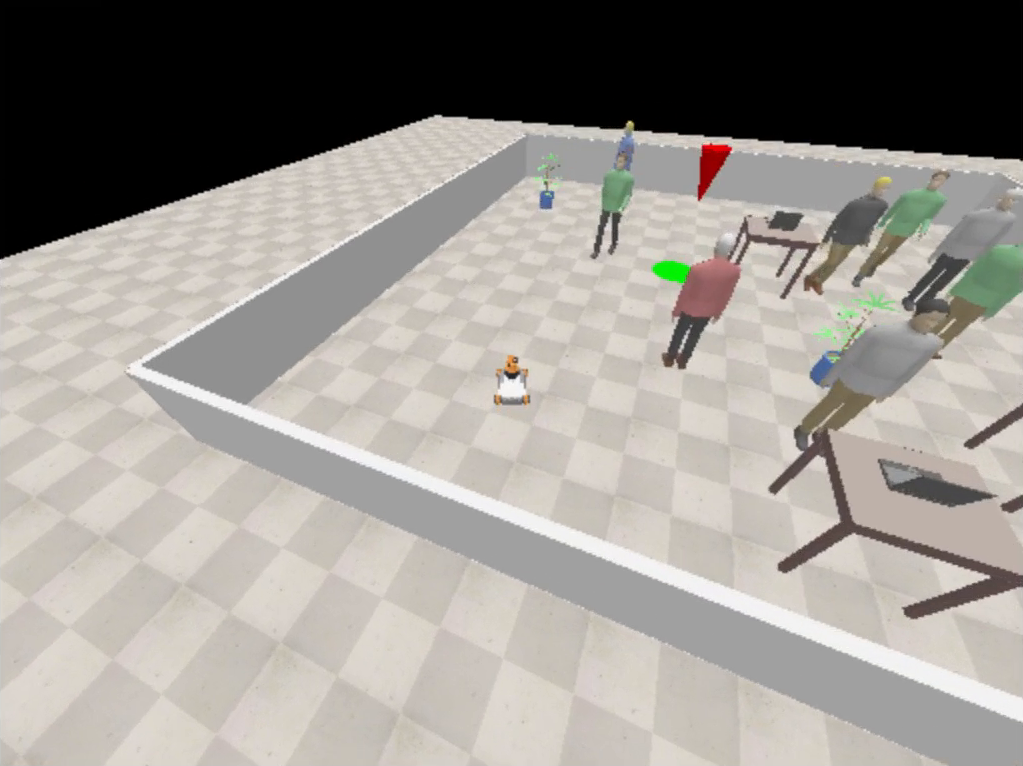}\label{fig:exp1_1}}
 \hspace{0.1cm}
\subfloat{\includegraphics[width=0.48\textwidth,keepaspectratio=true]{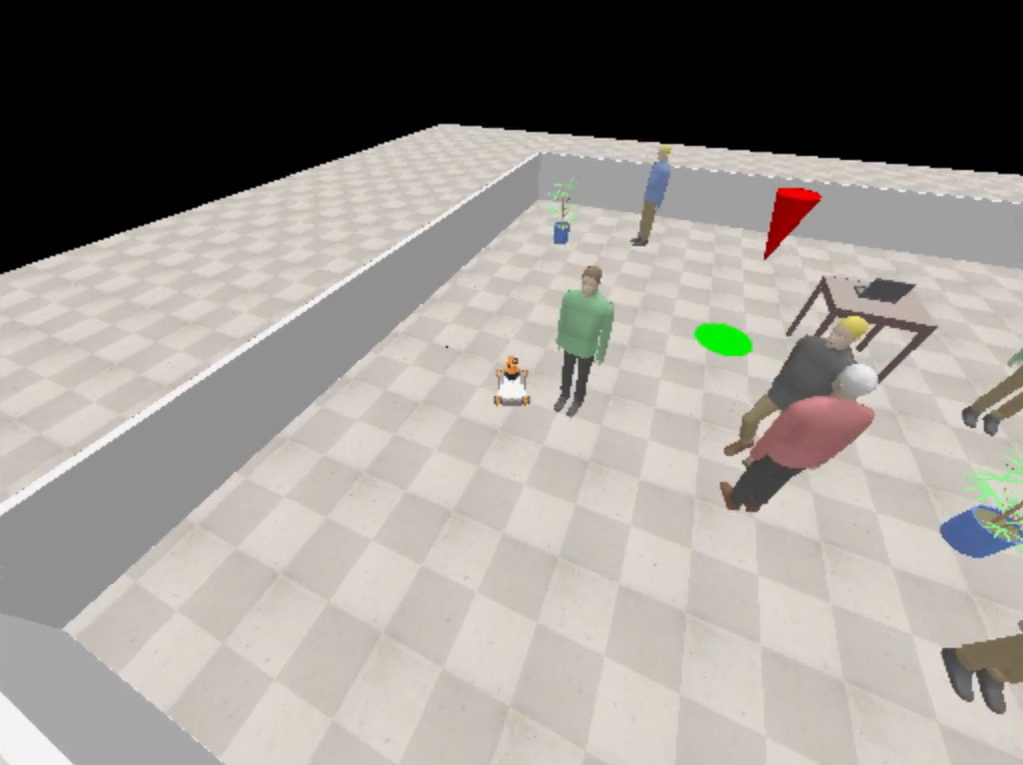}\label{fig:exp1_2}} \\
\subfloat{\includegraphics[width=0.48\textwidth,keepaspectratio=true]{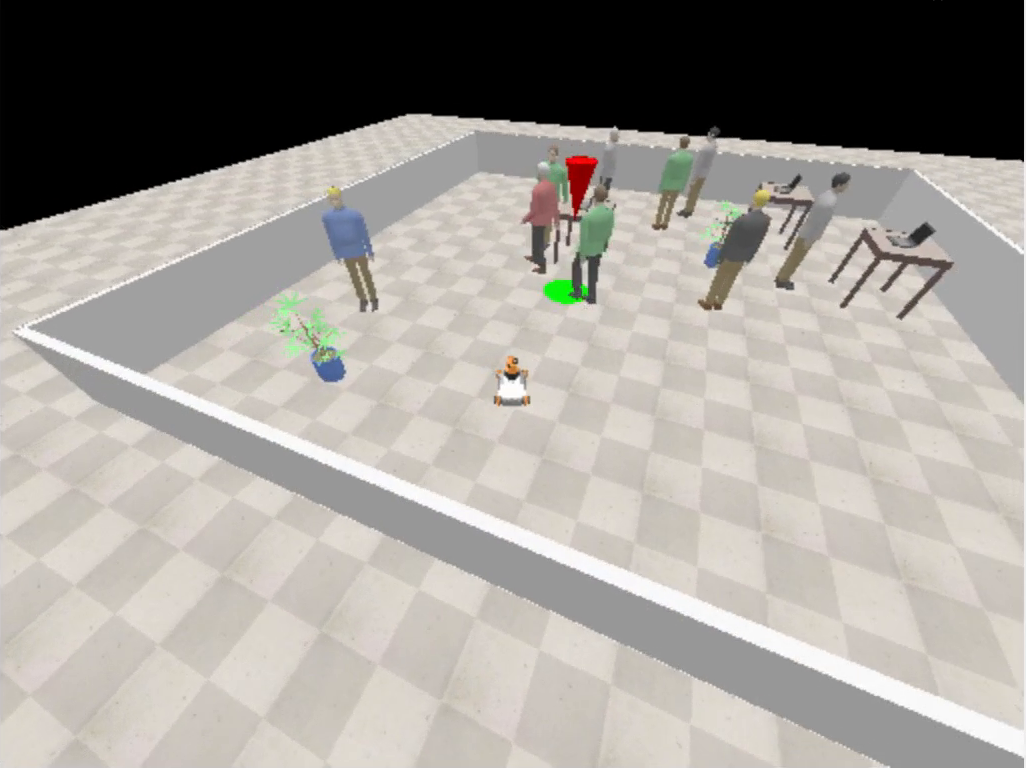}\label{fig:exp1_3}}
 \hspace{0.1cm}
\subfloat{\includegraphics[width=0.48\textwidth,keepaspectratio=true]{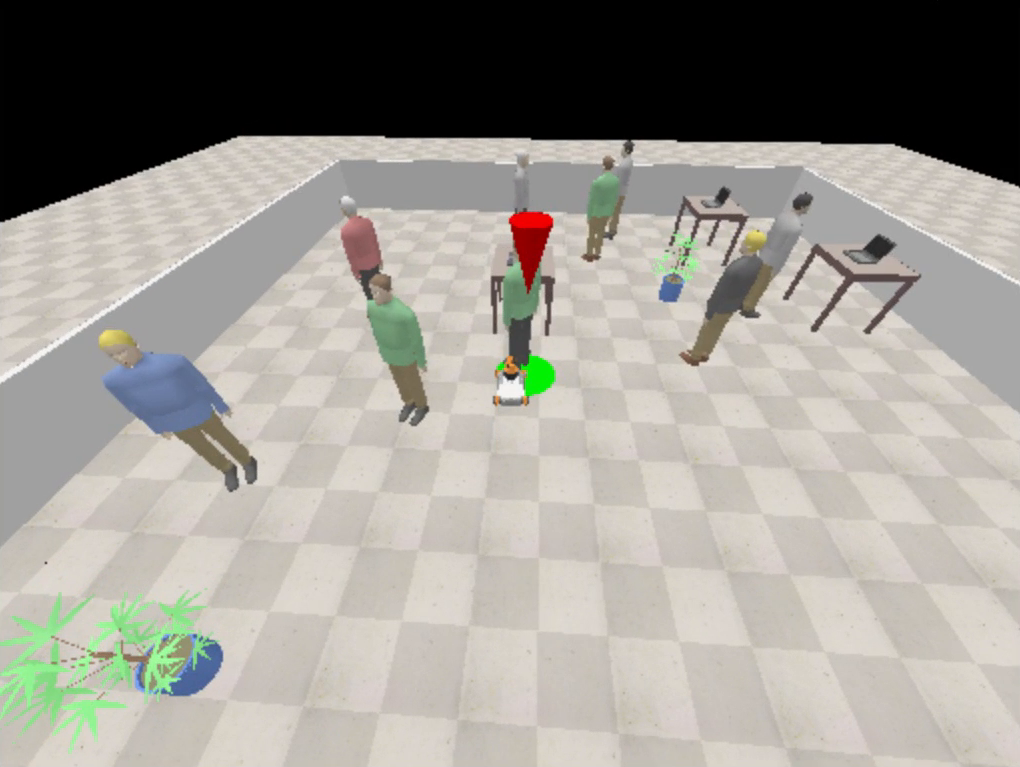}\label{fig:exp1_4}}
\caption{Execution of the robot using the trained model (second scenario). The robot avoids a human walking in the opposite direction but gets too close to a human when it reaches the goal.}
\label{fig:exp1}
\end{figure*}

\begin{figure*}[!ht] 
 \centering
\subfloat{\includegraphics[width=0.48\textwidth,keepaspectratio=true]{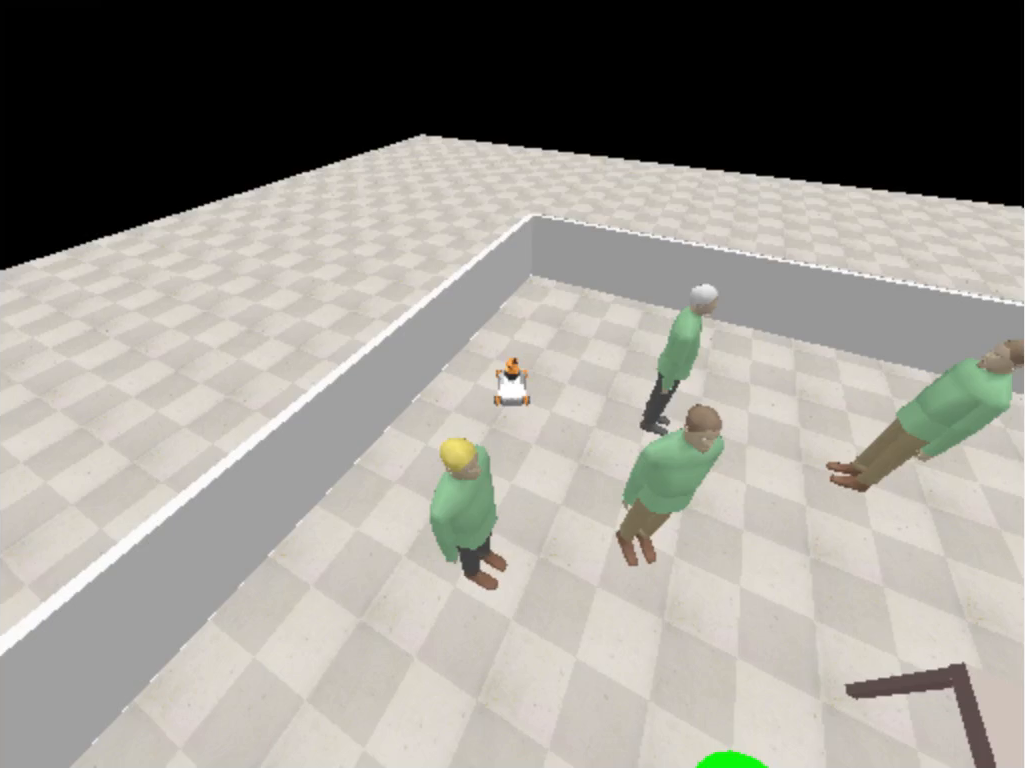}\label{fig:exp2_1}}
 \hspace{0.1cm}
\subfloat{\includegraphics[width=0.48\textwidth,keepaspectratio=true]{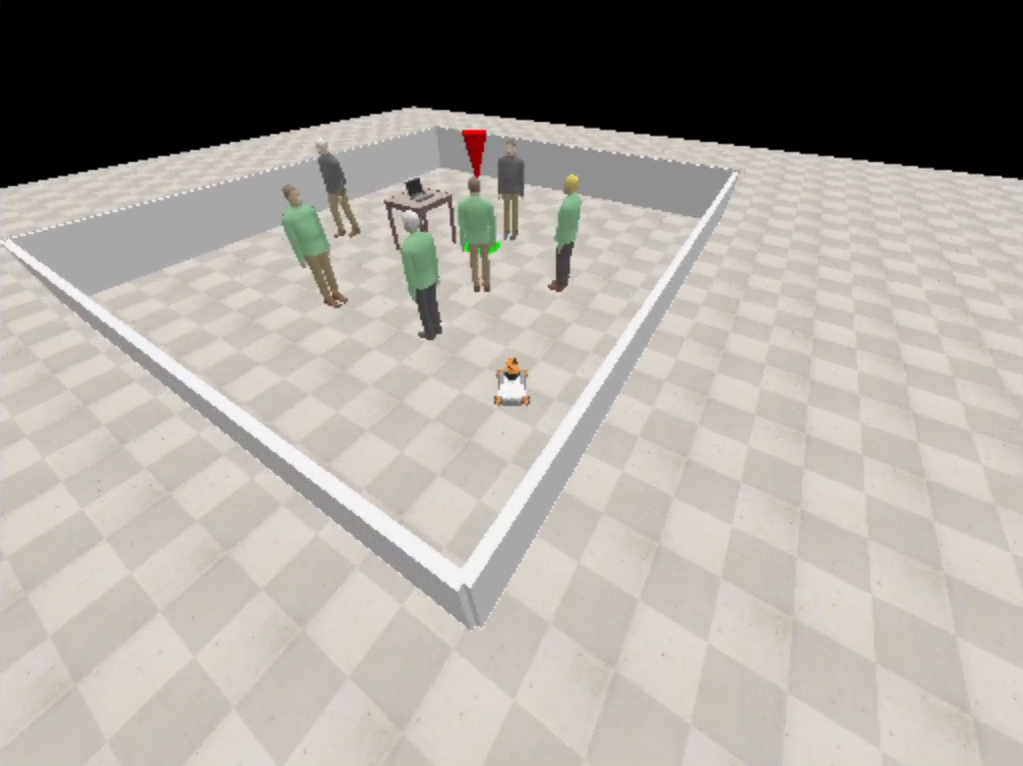}\label{fig:exp2_2}} \\
\subfloat{\includegraphics[width=0.48\textwidth,keepaspectratio=true]{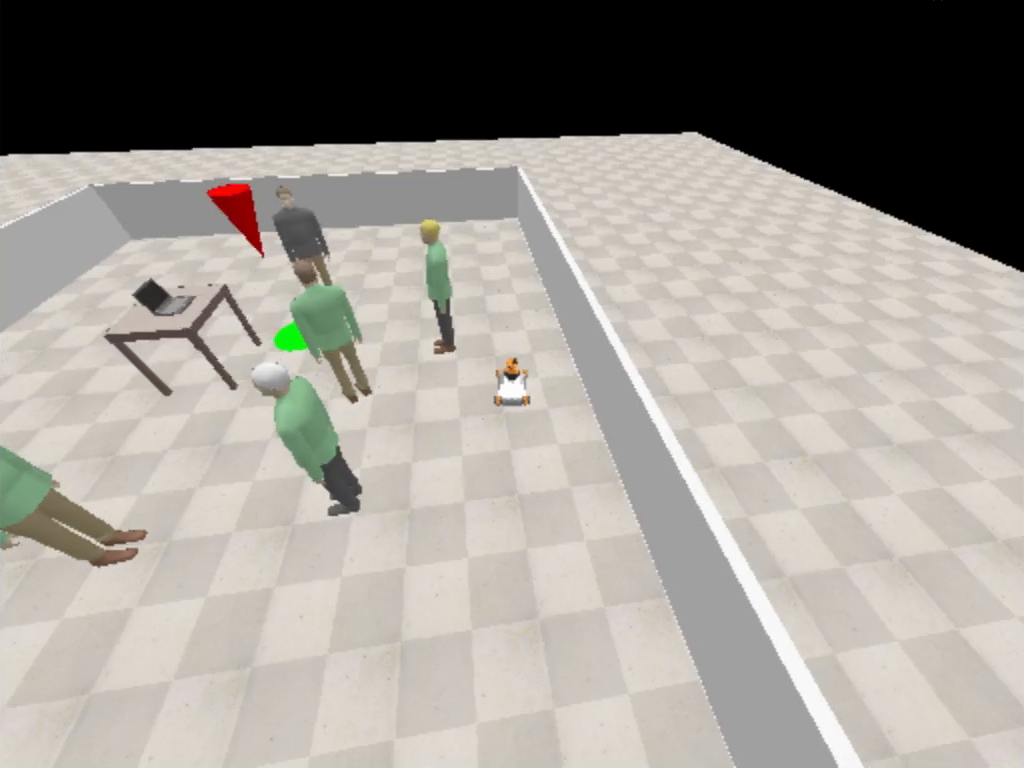}\label{fig:exp2_3}}
 \hspace{0.1cm}
\subfloat{\includegraphics[width=0.48\textwidth,keepaspectratio=true]{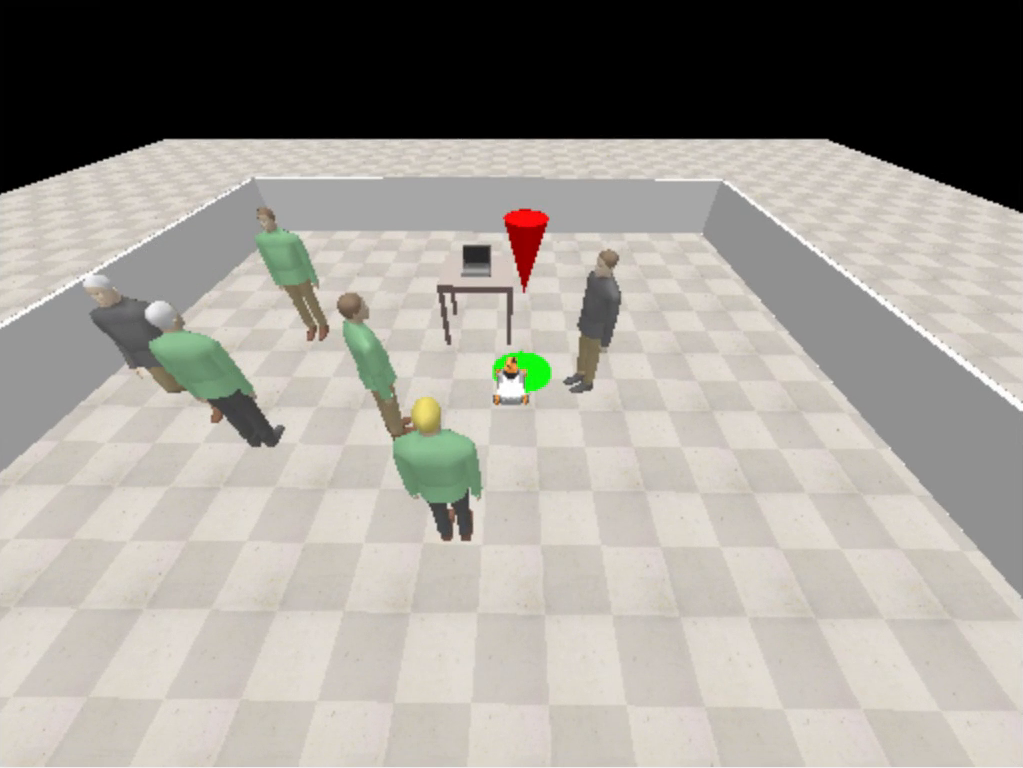}\label{fig:exp2_4}}
\caption{Execution of the robot using the trained model (third scenario). The robot changes its trajectory to avoid a group of humans on the way to the target. It gets into an interaction area to finally reach the goal position.}
\label{fig:exp2}
\end{figure*}

We obtained an MSE of $0.0230$ for the training set, and $0.0328$ and $0.0331$ for the development and test sets.
Results of the execution of the robot using the network output in different scenarios can be found in the following link: \url{http://shorturl.at/etLOW}.
Figures~\ref{fig:exp3}, \ref{fig:exp1} and~\ref{fig:exp2} show some snapshots of three of these executions.
Each sequence is ordered from top to bottom and left to right.
The model trained can reach the goal of $76.6\%$ of the times ($46$ out of $60$).
Given the limited number of samples and the complexity of some scenarios, this percentage can be considered reasonable.
Nevertheless, the behaviour of the robot is not always the most appropriate.
Sometimes the robot gets too close to humans and objects (Fig.~\ref{fig:exp1}).
Besides, in some scenarios, the robot interrupts people's interactions even though it would be possible to find an alternative path to reach the goal position (Fig.~\ref{fig:exp2}).
These results are in line with the literature, which does not recommend using supervised learning for robot control because the samples' distribution tends to be biased towards safe situations.
This bias makes robots unreliable in unexpected situations.
Besides showing an example of data gathering using the proposed toolkit, reproducing these results was one of the purposes of the use case.
The analysis of the influence of larger datasets in supervised learning and exploring alternative approaches to this problem, such as inverse reinforcement learning, is left for future research and lie out of the scope of the paper.

\section{Conclusions}
The experience working with the toolkit to gather data for the use case is deemed satisfactory, as $3092$ samples were gathered by only $6$ subjects who spent less than $31$ hours in total.
Although satisfactory, some limitations and desirable features that will be addressed in future works were highlighted:
\begin{itemize}
    \item Absence of a HUD-like display to provide information on top of the image. For instance, this could be used to point subjects to the goal's location when it is not visible.
    \item The meshes of the simulated humans sometimes intersect and go through walls that are meant to be solid.
    \item The orientation of the head of the simulated humans cannot be controlled. Controlling the head orientation could be useful to hint subjects about the future trajectory of the simulated humans.
\end{itemize}
\par

The first feature could be implemented by the user drawing on top of the image acquired from the camera. 
However, the API of SONATA will be extended to make the task easier.
\par

The humans' behaviour, which is related both to the collision detection and the lack of ability to move their heads, depends on how they are simulated in CoppeliaSim.
Enforcing collision detection could be solved by adding a control layer.
To solve the second issue, it would be necessary to implement a new controllable simulated human in CoppeliaSim as an extension of the current \textit{"path planning Bill"} model.
\par

As a final remark, it must be noted that any simulation will differ from real life to some extent.
Even if they are very realistic, subjects will know that they are in a simulation that might impact their behaviour.
\par

The software has been published in \url{https://github.com/ljmanso/sonata} under a permissive license.

%
%
\bibliographystyle{spmpsci_unsrt} 
\bibliography{bib}

\end{document}